\documentclass[conference]{IEEEtran}

\ifCLASSINFOpdf
  \usepackage[pdftex]{graphicx}
\else
  \usepackage[dvips]{graphicx}
\fi
\usepackage{pdfpages}
\usepackage{amsmath}
\usepackage{amssymb}
\usepackage{hyperref}
\usepackage{algorithm}
\usepackage{algpseudocode}
\usepackage{booktabs}
\usepackage{algorithmicx}
\usepackage{float}
\usepackage{subfigure}
\usepackage{soul}
\usepackage{enumitem}
\usepackage{adjustbox}

\newcommand{\policy}{\pi}

\newcommand{\params}{\theta}

\newcommand{\pparams}{{\phi}}

\newcommand{\Q}{Q}

\usepackage[T1]{fontenc}

\begin{document}
\title{Heuristic Algorithm-based Action Masking Reinforcement Learning (HAAM-RL) with Ensemble Inference Method}
 
\author{\IEEEauthorblockN{Kyuwon Choi\IEEEauthorrefmark{1}}
\IEEEauthorblockA{AgileSoDA \\
Seoul, South Korea\\
kwchoi@agilesoda.ai}
\and
\IEEEauthorblockN{Cheolkyun Rho\IEEEauthorrefmark{1}}
\IEEEauthorblockA{AgileSoDA \\
Seoul, South Korea\\
cheolkyun.rho1992@agilesoda.com}
\and
\IEEEauthorblockN{Taeyoun Kim\IEEEauthorrefmark{1}}
\IEEEauthorblockA{AgileSoDA \\
Seoul, South Korea\\
klumblr@agileoda.ai}
\and
\IEEEauthorblockN{Daewoo Choi\IEEEauthorrefmark{2}}
\IEEEauthorblockA{Hankuk University of Foreign Studies \\
Yongin, South Korea\\
daewoo.choi@hufs.ac.kr}
}

\maketitle
\begin{abstract}
This paper presents a novel reinforcement learning (RL) approach called HAAM-RL (Heuristic Algorithm-based Action Masking Reinforcement Learning) for optimizing the color batching re-sequencing problem in automobile painting processes. The existing heuristic algorithms have limitations in adequately reflecting real-world constraints and accurately predicting logistics performance. Our methodology incorporates several key techniques including a tailored Markov Decision Process (MDP) formulation, reward setting including Potential-Based Reward Shaping, action masking using heuristic algorithms (HAAM-RL), and an ensemble inference method that combines multiple RL models. The RL agent is trained and evaluated using FlexSim, a commercial 3D simulation software, integrated with our RL MLOps platform BakingSoDA. Experimental results across 30 scenarios demonstrate that HAAM-RL with an ensemble inference method achieves a 16.25\% performance improvement over the conventional heuristic algorithm, with stable and consistent results. The proposed approach exhibits superior performance and generalization capability, indicating its effectiveness in optimizing complex manufacturing processes. The study also discusses future research directions, including alternative state representations, incorporating model-based RL methods, and integrating additional real-world constraints.
\end{abstract}


\IEEEpeerreviewmaketitle

\section{Introduction}
After the emergence of Alpha-Go in 2016, numerous companies including DeepMind \cite{Silver2016} showed positive potential in the field of reinforcement learning (RL). However, in practical industry implementations, the effective utilization of RL techniques has been notably scarce \cite{schoettler2019deep}. The principal challenge stems from the fact that in RL, there isn't a universally agreed-upon method for precisely defining a Markov Decision Process (MDP), which is crucial for implementing RL algorithms. The expertise and practical experience of the practitioner greatly influence model training performance. The iterative process of trial and error, while essential, can extend training times and impede convergence. Achieving a balance between exploration and optimization is crucial for efficient and effective model training \cite{Dulac-Arnold2021}, \cite{rlblogpost}. This naturally leads to the industry veering off to looking at effective heuristic algorithms to optimize its decision process \cite{kothapalli2023learningbased}. The entire automobile production process is highly complicated and is mostly run by manual and human decision-based \cite{su14042408}, \cite{guo2019hybrid}, \cite{EvolutionOfCognitiveDemand}, \cite{doi:10.1080/21693277.2020.1737592}. However, with the diverse need among automobile customers increasing, coupled with the trend of complex decision-making factors due to multi-variant, small-batch production, challenges are on the rise. In this paper, we combined heuristic algorithms with RL to show practical, yet successful application in the painting process of the automobile production process. Our RL algorithm is Soft Actor-Critic (SAC) \cite{haarnoja2018soft} based with incorporation of action masking through heuristic algorithms and an ensemble inference method; to verify our algorithm, we compared it in 30 different scenarios and saw 16.25\% increase in performance.

The rest of the paper is organized as following: Section~\ref{sec:rel} provides an overview of related research addressing the identified issue. Section~\ref{sec:meth} introduces the main contributions of this paper. Section~\ref{sec:algo} explains our main RL algorithm, HAAM-RL, in more detail. Section~\ref{sec:exp} furnishes the experimental setup for validation purposes. Section~\ref{sec:result} explains the result of the  experiments to verify our system. Lastly, Section ~\ref{sec:con} encapsulates the the discussion and conclusion pertaining to this problem.

\begin{figure}[htbp]
    \centerline{\includegraphics[width=9cm]{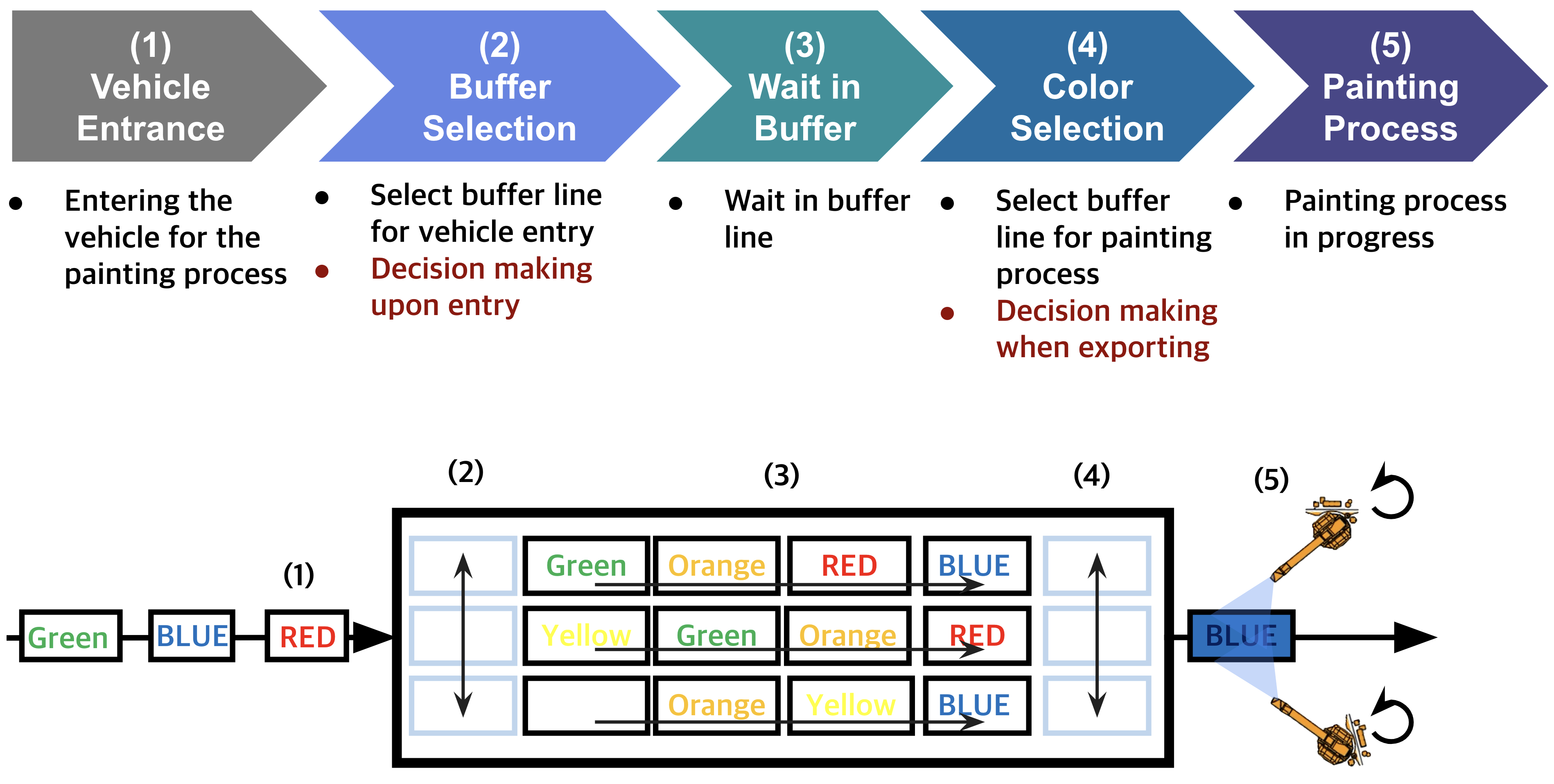}}
    \caption{Abstract representation of the painting process in the automobile production line}
    \label{fig:FIG1}
\end{figure} 

\section{Related Work}
\label{sec:rel}
In the manufacturing and logistics industries, heuristic-based approaches have been used to manage the diverse and complex processes \cite{li2021learning}. Many research have been conducted on the sequencing problem of multi-version products to ensure production flexibility. However, existing heuristic methods have limitations in adequately reflecting real-world situations or in accurately predicting logistics when algorithms are modified \cite{Meli_2024}. Several factors including the impractical time complexity of execution and the misrepresentation of real-world settings hinder the effective utilization of heuristic algorithms or result in sub-optimal performance. Moreover, real-life constraints, such as the costs associated with equipment changes, further impede their application. 

\subsection{Color-batching Re-sequencing Problem}
The problem can be restated as a Color-batching Re-sequencing Problem (CRP), where the objective is to minimize the number of color change of the paint during the car painting process, as shown in Figure \ref{fig:FIG1}. There have been numerous new attempts to optimize CRP. \cite{10.1007/978-3-030-13273-6_15} utilized Follow-up Production Control (FPC) ideology to suggest Follow-up Sequencing Algorithm (FuSA) to be applied. In this research, efforts were made to enhance the flexibility of the painting process by considering not only conveyor systems but also utilizing buffers to manage the production flow. Furthermore, the authors conducted research aiming to validate the performance through various experiments and minimize discrepancies with real-world scenarios \cite{10.1007/978-3-319-97490-3_36}. However, it was necessary to consider additional technical constraints such as periodic cleaning of the equipment and the cost of color changes \cite{huang2022paint}. Consequently, advanced concepts like Car Sequencing Problem 4.0 (CSP 4.0) were proposed, leading to research on optimizing painting processes that take into the aforementioned constraints \cite{10.1007/978-3-030-13273-6_15}. Various research endeavors have attempted to optimize manufacturing logistics processes through numerous approaches. However, these efforts have often been unsuccessful due to inadequate consideration of real-world complex constraints or an inability to achieve high verification efficiency. 

\subsection{Deep Reinforcement Learning}
With the increase in the technology of RL, it is being applied in many applications, including manufacturing, logistics industry \cite{stranieri2023comparing}. RL is expressed in terms of Markov Decision Process (MDP), as defined in \cite{puterman1994markov}; MDP is consisted of $<\mathcal{S},\mathcal{A},\mathcal{P},\mathcal{R},\gamma>$. The main goal of deep RL is for the agent to learn a policy, $\mathcal{P}$, that will maximize the reward, $\mathcal{R}$, in a given state, $\mathcal{S}$. Therefore, the performance varies significantly depending on how rewards are defined \cite{NEURIPS2022_6255f223}. Various techniques are introduced to automate reward definitions \cite{9558764} or assign rewards based on state information \cite{10099211}. There is also research on defining reward functions related to inverse RL, which is implicitly inferring reward functions based on the agent's actions \cite{10.5555/3295222.3295421}. 

There also exists research on application of ensemble of neural networks in deep RL for different purposes \cite{jedrzejowicz2020deep}, \cite{sheikh2022dns}, \cite{chen2021randomized}. These methods have proven to increase efficiency and performance. 

We utilized RL to solve CRP at a practical scale that pre-existing heuristic algorithms weren't able to do. We incorporated Potential-Based Reward Shaping (PBRS) \cite{10099211} to increase the efficiency during the training process, and applied the ensemble methodology for the stability of the RL agent and for the increase in performance in general situations. This agent was trained and verified on FlexSim, a 3D Simulation Modeling and Analysis Software \cite{FlexSim}. 

\section{Methodology}
\label{sec:meth}
Our main contribution points can be categorized as three key components: 1) the development of a novel RL MDP, 2) the utilization of ensemble method involving multiple RL models during evaluation, and 3) connecting FlexSim with our proprietary RL MLOps platform, BakingSoDA (BASD) \cite{agilesoda2021}.

\subsection{MDP formulation}
\begin{figure}[h!]

    \centerline{\includegraphics[width=8cm]{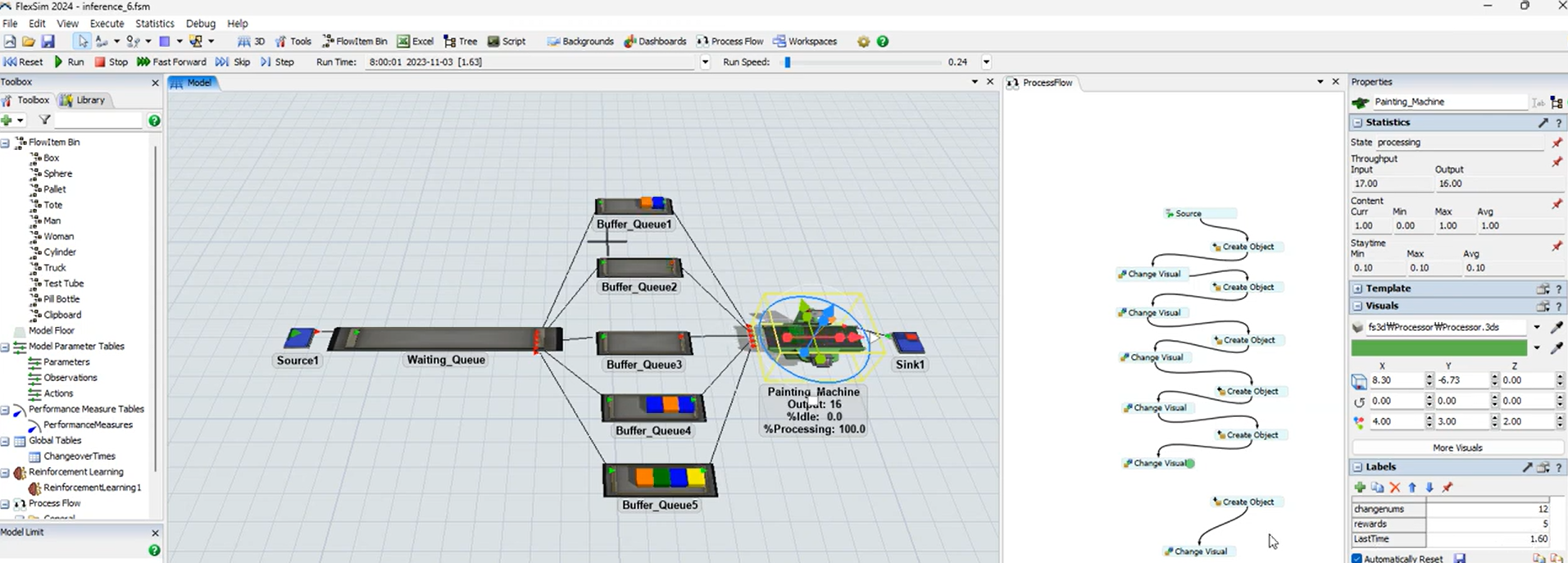}}
    \caption{Our FlexSim simulator setup}
    \label{fig:FIG2}
    
\end{figure} 

We set up a FlexSim simulator as Figure \ref{fig:FIG2}. With one input queue, 5 buffers of 5 slots per buffer, and one output queue, we concatenate each index into a 2D numpy array with the length of 27. This is defined as a state $\mathcal{S}$ that the agent receives from the environment during the training process. The action, $\mathcal{A}$, is denoted as a pair-wise action space of 25, a combination of 5 selections of the input to the buffer and 5 selections of the output from the buffer. We implemented action masking methodology to mask out the invalid actions to reduce the number of valid actions that the agent needs to select from. For example, if the first buffer is full with 5 cars, the agent cannot select an action that has a first input buffer. We also implemented preexisting heuristic algorithms aforementioned that acts as a masking algorithm to expedite the training process. As a reward model, we incorporated PBRS \cite{10099211} to calculate the potential from the state, penalty when the final color changes, and positive reward when the current color of the car is the same as that of the previous car with more positive reward for the number of consecutive times the color hasn't changed.

\begin{itemize}
    \item $\mathcal{S}$: State space $\in \mathbb{R}^{27}$. A 2D array of all slots of the system 
    \item $\mathcal{A}$: Discrete action space $\in \mathbb{R}^{25}$. 5 input $\times$ 5 output
    \item $\mathcal{R}$: Reward function $R = P + C$.
        \begin{itemize}
            \item Potential ($P$): Potential Based Reward Shaping
            \item Car reward ($C$): Either penalty or reward depending on the current and previous color
        \end{itemize}
\end{itemize}

\begin{equation}
    \text{Car reward} = 
    \begin{cases}
        -0.1 & \text{if } \text{prev color} \neq \text{cur color} \\
        0.1 + \text{dup} & \text{otherwise}
    \end{cases}
\end{equation}
Dup is defined as the number of consecutive times the specific color has been chosen.

\subsection{Heuristic Algorithm-based Action Masking}
\begin{figure}[h!]
    \centerline{\includegraphics[width=9cm]{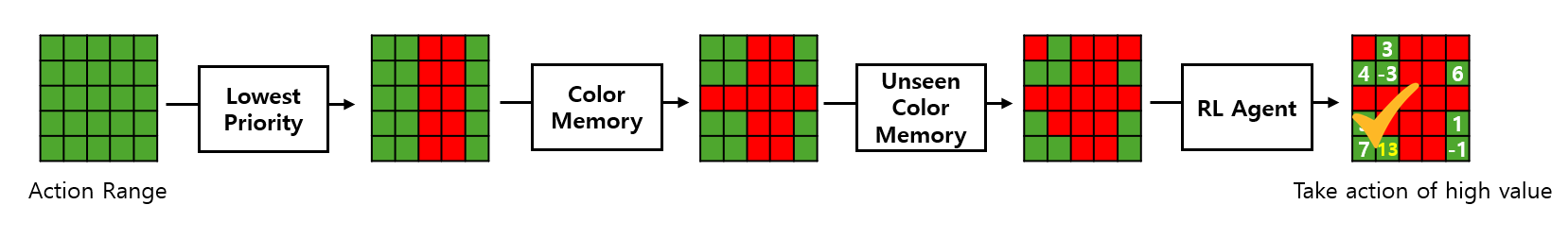}}
    \caption{An example of heuristic algorithm-based action masking technique (Green: valid action, Red: invalid action)}
    \label{fig:FIG3}
\end{figure}

Many approaches have been made to modify the MDP to the vanilla RL algorithm at the beginning, but weren't able to produce satisfying results. Our analysis suggests that a prevalent factor contributing to this poor convergence of the RL agent is the substantial discrepancy in dimensions between the action and state spaces, as the action dimensions is relatively larger for this state dimension. There is a need to reduce the action dimension to achieve stability and faster convergence in training. Therefore, we have adopted a way to employ action masking using heuristic algorithms. This methodology has been proved to improve performance in other RL papers \cite{6502216}, \cite{krasowski2023provably}, \cite{cappart2020combining}, \cite{cheng2021heuristicguided}. We have researched and implemented different heuristic algorithms (LP, CM \cite{10.1007/978-3-319-97490-3_36}, and our own UCM) regarding this problem and Figure \ref{fig:FIG3} represents the final way we have adopted to reduce the action space through action masking. This resulted in the stability and faster convergence of the RL agent.

\begin{figure}[h!]
    \centerline{\includegraphics[width=8cm]{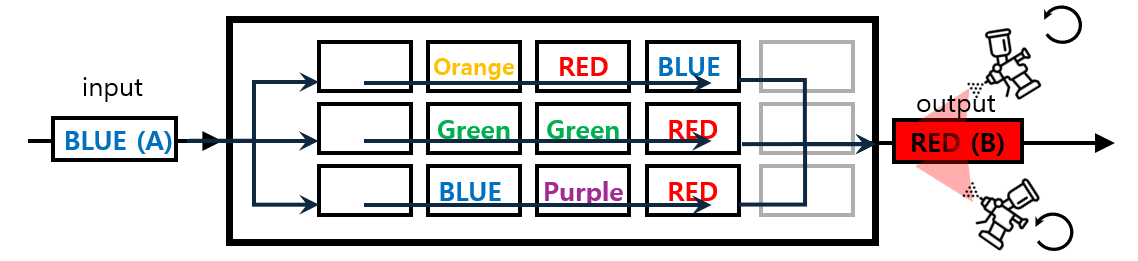}}
    \caption{An example of heuristic algorithm used in heuristic algorithm-based action masking}
    \label{fig:FIG4}
\end{figure}

The detailed explanation of different heuristic algorithms used in this paper is shown below and in Figure \ref{fig:FIG4}. 

\begin{itemize}
    \item Lowest Priority (LP) : Applied only on the input of the buffers. Insert into the buffer with the least diverse color.
    \item Color Memory (CM) : Applied both on the input and output of the buffers. Insert into the buffer where the vehicle's color is the same as the last vehicle in the buffer.
    \item Unseen Color Memory (UCM) : Applied only on the output of the buffers. Extract the vehicle with the most colors when there is no vehicle of the same color as the last extracted vehicle's color.
\end{itemize}

If there still exists a situation where two buffers are the same after applying those algorithms, the most top buffer is chosen.

\subsection{Ensemble Inference Method}
During training with variations of the neural network hyper-parameters and reward tuning, we were able to collect numerous RL models. At first, each RL model was deployed independently during the inference process to see individual performance. There have been instances where one model exhibited poorer overall performance compared to another model in terms of average metrics; however, it demonstrated superior performance in specific scenarios. Therefore, we used two different methods to utilize these RL models when evaluating using FlexSim: 1) hard voting and 2) soft voting. Hard voting is defined as the selection of most frequently chosen action across all models within a given state. However, our approach deviates from strict hard voting. Instead, we calculate the number of times of the most selected action and compare it against a predefined threshold value, which is delineated by Equation \ref{eq:hardvote}. 

\begin{equation}
    \text{count} = \frac{\text{len(models)}}{2} - 1
\label{eq:hardvote}
\end{equation}

\begin{align}
\text{ensemble method} = \begin{cases}
h.v. & \text{val $>$ count}\\
s.v. & \text{otherwise}.
\end{cases}
\label{eq: ensem}
\end{align}

In terms of the notation, h.v. denotes hard voting, s.v. denotes soft voting, and val is the number of times of the most selected action from all models. To explain in more detail, if there are 10 models, the hard voting mechanism will only be invoked if the number of most frequent action is greater than or equal to 4. If not, the soft voting mechanism will be called. 

The soft voting is normalizing each logits value from the distribution from a model and summing for all the index within the action space throughout all the models. Soft voting can be expressed as following: Let $M$ be the number of models and $N$ be the number of elements in the entire action list. Let $\text{logit}_{ij}$ represent the logit value for the $i$-th model and the $j$-th element, where $i \in \{1, 2, \ldots, M\}$ and $j \in \{1, 2, \ldots, N\}$.

Then, after normalizing each logit value, the summation for the $j$-th element can be expressed as:

\begin{equation}
\text{Sum}_{j} = \sum_{i=1}^{M} \text{logit}_{ij}
\label{eq:softvoting}
\end{equation}

Then, the agent will choose the maximum value from all indices as its final action.
These two methods account for all the models for a given state to choose the best possible action. Detailed explanation is denoted in Algorithm \ref{alg:Ensemble}

\subsection{Training using MLOps Platform for RL with FlexSim}
AgileSoDA has its own RL MLOps Platform called BakingSoDA \cite{agilesoda2021}. BakingSoDA supports UI-based interface algorithms for analysts and developers who are not familiar with RL, the development and operation of RL models that self-assess performance, and neural network hyper-parameter auto tuning function. BakingSoDA provides an internal simulator functionality tailored to the user's business characteristics, where the user can develop and customize own RL environment. However, for this research, we used an external Simulator functionality. 

\begin{itemize}
    \item Internal Simulator Functionality : Through the provided Simulator Server, the user customizes the template to create RL environment for training that is suitable to BakingSoDA's framework.
    \item External Simulator Functionality : Integration between a commercial simulator and BakingSoDA is possible through data exchange with BakingSoDA via HTTP communication. Furthermore, the Connector REST API, defined according to specifications, should be utilized for BakingSoDA to make calls to the simulator
\end{itemize}

\begin{figure}[h!]
    \centerline{\includegraphics[width=9cm]{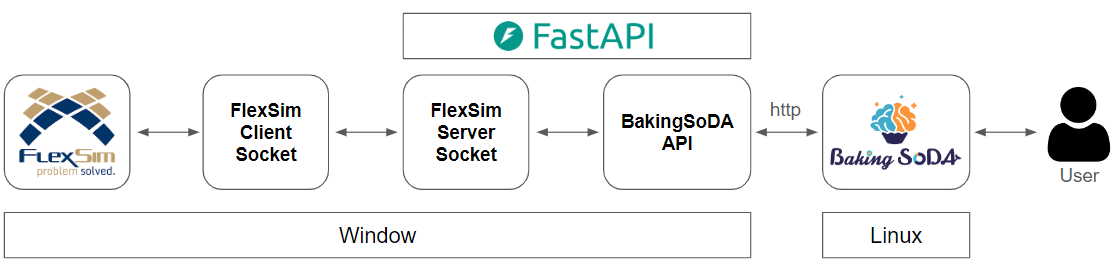}}
    \caption{Process of integration FlexSim with BakingSoDA API}
    \label{fig:FIG5}
\end{figure}

This paper utilizes FlexSim, a simulation software platform mainly used for modeling and simulating complex systems across various industries such as manufacturing and logistics. It allows users to create dynamic models of their processes, test different scenarios, and analyze the performance of their systems. It also provides a visual environment where users can build 3D models and visualize the behavior of their systems in real-time \cite{FlexSim}.
We successfully created our own FlexSim environment as shown in Figure \ref{fig:FIG2} connected it through external API to facilitate training and inference stages of RL as shown in Figure \ref{fig:FIG5}. 

\section{HAAM-RL}
\label{sec:algo}
\begin{figure}[h!]
    \centerline{\includegraphics[width=9cm]{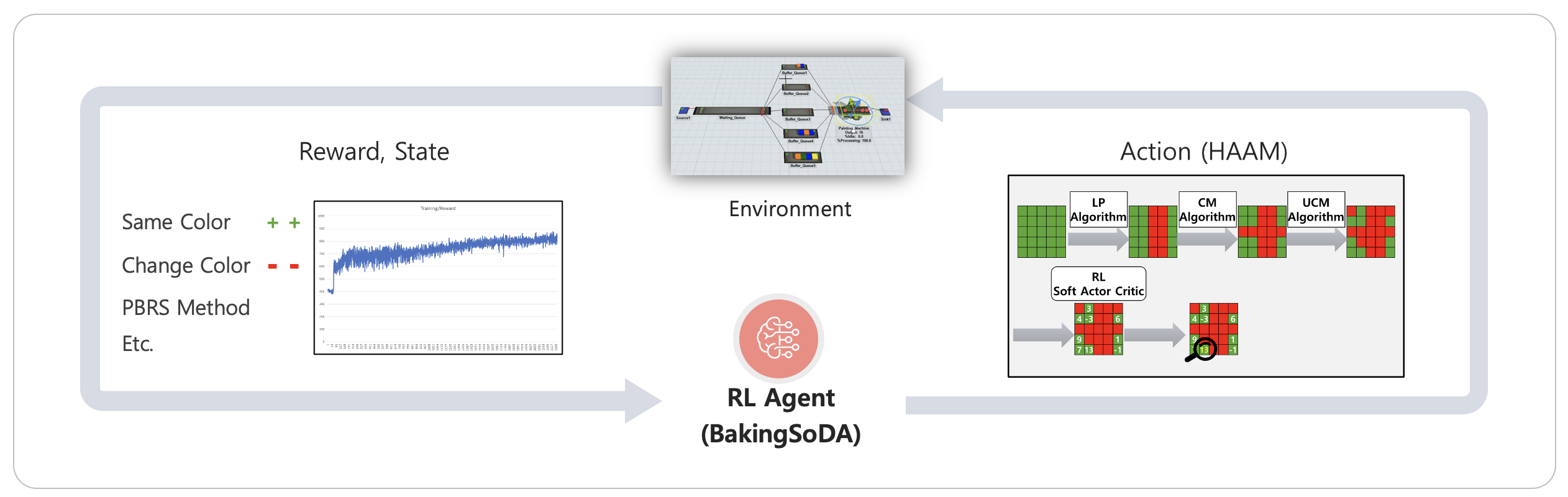}}
    \caption{HAAM-RL framework}
    \label{fig:FIG6}
\end{figure}

As mentioned in Section 3, to solve this problem, we have 1) defined our own MDP, 2) masked action using heuristic algorithms, 3) applied ensemble inference method, and 4) integrated BakingSoDA with FlexSim to perform RL. Overall, HAAM-RL's architecture is shown in Figure \ref{fig:FIG6}. 

Out of multiple algorithms in RL, we employ SAC \cite{haarnoja2018soft} algorithm for our algorithm as SAC inherits the concept of maximizing entropy and expected reward. We slightly modify SAC in order to incorporate heuristic algorithm masking methodology. To explain once again, upon receiving state information from the environment, the agent calculates the valid action list from heuristic algorithms. Then, other invalid actions will be masked and the agent chooses an action from the valid action list. This reduces the number of choices the agent has to choose from and will facilitate faster and stable training. We present HAAM-RL pseudo Code in Algorithm \ref{alg:HAAM-RL}. Also, we present ensemble inference method, explained in Section \ref{sec:meth}, pseudo code in Algorithm \ref{alg:Ensemble}. 

\begin{algorithm}[htbp]
\caption{HAAM-RL}
    \label{alg:HAAM-RL}
\begin{algorithmic}[1]
\State Initialize policy network $\pi_\phi$, Q-value networks $Q_{\theta_1}$, $Q_{\theta_2}$, replay buffer $\mathcal{D}$, and masking vector $\mathbf{m}$ with ones
\State Initialize target networks $\bar{\pi}_{\bar{\phi}} \gets \pi_\phi$, $\bar{Q}_{\bar{\theta}_1} \gets Q_{\theta_1}$, $\bar{Q}_{\bar{\theta}_2} \gets Q_{\theta_2}$
\For{each iteration}
   \For{each environment step}
       \State Receive state $s_t$ from the FlexSim environment
       \State Calculate heuristic algorithms (LP, CM, UCM) to get candidate action list $\mathcal{A}_t$
       \State Update masking vector $\mathbf{m}$ by setting entries for actions not in $\mathcal{A}_t$ to 0
       \State Sample action $a_t \sim \pi_\phi(\cdot|s_t) \odot \mathbf{m}$ from the masked policy
       \State Take action $a_t$ and observe next state $s_{t+1}$ from FlexSim
       \State Calculate reward $r_t$
       \State Store $(s_t, a_t, r_t, s_{t+1})$ in replay buffer $\mathcal{D}$
   \EndFor
   \For{each gradient step}
        \State $\params_i \leftarrow \params_i - \lambda_Q \hat{\nabla}_{\params_i} J_\Q(\params_i)$ for $i\in\{1, 2\}$ \Comment{Update the Q-function parameters}
        \State $\pparams \leftarrow \pparams - \lambda_\policy \hat{\nabla}_\pparams J_\policy(\pparams)$\Comment{Update policy weights}
        \State $\alpha \leftarrow \alpha - \lambda \hat{\nabla}_\alpha J(\alpha)$ \Comment{Adjust temperature}
        \State $\bar{\params}_i\leftarrow \tau \params_i + (1-\tau)\bar{\params}_i$ for $i\in\{1,2\}$\Comment{Update target network weights}
    \EndFor
\EndFor
\end{algorithmic}
\end{algorithm}

\begin{algorithm}[htbp] 
\caption{Ensemble Inference Method}
    \label{alg:Ensemble}
\begin{algorithmic}[1]
\Require Trained policy networks $\{\pi_{\phi_i}\}_{i=1}^N$, heuristic algorithms (LP, CM, UCM)
\Require Hard voting threshold $\tau_h$, soft voting weights $\{\alpha_i\}_{i=1}^N$
\State Receive initial state $s_0$ from the environment
\For{each time step $t$}
   \State Receive state $s_t$ from the FlexSim environment
   \State Calculate heuristic algorithms to get candidate action list $\mathcal{A}_t$
   \State Update masking vector $\mathbf{m}$ by setting entries for actions not in $\mathcal{A}_t$ to 0
   \For{each policy network $\pi_{\phi_i}$}
       \State Sample action $a_t^i \sim \pi_{\phi_i}(\cdot|s_t) \odot \mathbf{m}$ from the masked policy
    \EndFor
   \If{\# of most frequent action $> \tau_h$}
       \State use hard voting
   \Else
       \State use soft voting
   \EndIf
   \If{hard voting}
       \State $a_t \gets \text{argmax}_a \sum_{i=1}^N \mathbb{I}(a_t^i = a)$ \Comment{Choose most frequent action}
   \ElsIf{soft voting}
       \State $a_t \gets \text{argmax}_a \sum_{i=1}^N \alpha_i \pi_{\phi_i}(a|s_t)$ \Comment{Weighted average}
   \EndIf
   \State Take action $a_t$ and observe next state $s_{t+1}$ from FlexSim
   \State $s_t \gets s_{t+1}$
\EndFor
\end{algorithmic}
\end{algorithm}

\section{Experiment Details}
\begin{figure}[htbp]
    \centerline{\includegraphics[scale=0.18]{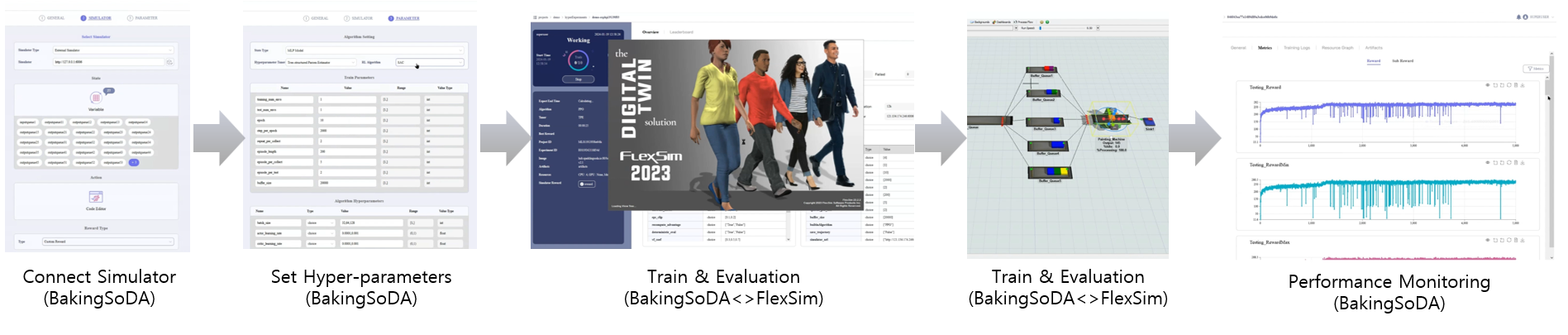}}
    \caption{Flow of Training \& Evaluation using BakingSoDA \& FlexSim}
    \label{fig:FIG7}
\end{figure}

In this section, we evaluate HAAM-RL to verify its effectiveness. We 1) used the same item distribution as \cite{10.1007/978-3-319-97490-3_36}, 2) created 30 different scenarios using the distribution to mitigate randomness, and 3) incorporated FlexSim to calculate the metric used for comparison.
\begin{itemize}
    \item Dataset distribution : Color 1: 6\% Color 2: 38\% Color 3: 29\% Color 4: 14\% Color 5: 10\% Color 6: 3\% \cite{10.1007/978-3-319-97490-3_36}
    \item Evaluation scenario : Based on the distribution, created 30 scenarios of 100 vehicles. Each scenario was tested using the heuristic algorithm \cite{10.1007/978-3-319-97490-3_36} and HAAM-RL. The total number of times the color changed was used as the metric for comparison.  
    \item Evaluation method : Used FlexSim to conduct the experiment as shown in Figure \ref{fig:FIG7}.
\end{itemize}

Additionally, we conducted approximately 100 experiments and will now provide insights into experiments exhibiting poor performance. We developed a RL environment with a progressive increase in complexity. Initially, the environment featured an even distribution of colors, then this distribution changed during the training process. Despite changing the distribution, we found that it did not have a significant impact on performance. As reward engineering is important for RL training, different values of the constants in the reward equation were used, but found to be insignificant. Results showed that integrating PBRS to the existing reward function was effective and gave good performance. Neural network's hyper-parameter tuning did not yield substantial differences in performance; final hyper-parameters used for training are shown in Table \ref{tab:hyperparams}.

\label{sec:exp}
\begin{table}[htbp]
\centering
\caption{Hyper-parameter Configuration}
\label{tab:hyperparams}
\begin{tabular}{p{4cm}p{4cm}}
\toprule
\textbf{Hyper-parameter} & \textbf{Value} \\
\midrule
Epoch & 10000 \\
Steps per epoch & 200 \\
Step per collect & 1 \\
Test num &1 \\
Batch size & 16\\
Episode per collect & 16 \\
Episode per test & 10 \\
Activation & ReLu \\
Hidden sizes & [64,64] \\
Norm layer & False \\
Num atoms & 1 \\
Use Dueling & False \\
Optimizer & Adam \\
Actor learning rate & 0.0001 \\
Critic learning rate & 0.0001 \\
Alpha learning rate & 0.0001 \\
buffer size & 20000 \\
discount factor & 0.99 \\
Reward normalization & False \\
Action scaling & False \\
$\tau$ & 0.005 \\
Estimation step & 3 \\
$\alpha$ & 0.05 \\
Deterministic eval & True \\
\bottomrule
\end{tabular}
\end{table}

\section{Results}
\label{sec:result}
Experiments show that our final algorithm, HAAM-RL showed 16.25\% improvement compared to the heuristic algorithm. The heuristic algorithm required 34 color changes for 100 vehicles, whereas HAAM-RL exhibited a lower number of color changes at 29. As the number of vehicles and the complexity of the environment increases, we expect HAAM-RL to decrease the overall cost while increasing efficiency in productions.  More detailed explanation can be found in Appendix \ref{sec:appendix_exp} of our experiment.

Also, to verify the result's stability and generalization potential, we also conducted experiments on 30 scenarios and analyzed their variance and standard deviation. The results showed a mean of 29.57, a variance of 6.530, and a standard deviation of 2.555. Furthermore, a 1 sample t-test was 0.05, which shows the minimal fluctuation in data. Through this analysis, we verified the stability of the proposed algorithm's performance. 

\begin{table}[htbp]
\centering
\caption{Experiment Results}
\label{tab:exp}
\begin{tabular}{|c|c|c|}
\hline
\textbf{Exp. \#} & \textbf{HAAM-RL} & \textbf{Note}\\
\hline
1 & -10.78\% & Initial Exp \\
\hline
2 & -1.58\% & State \& Reward Tuning \\
\hline
3 & 1.12\% & Hyper-parameter Tuning \\
\hline
4 & 5.14\% & Masking Method \\
\hline
5 & 12.23\% & Ensemble Method \#1\\
\hline
6 & 15.24\% & Ensemble Method \#2\\
\hline
7 & 16.28\% & Best Model \\
\hline
\end{tabular}
\end{table}

Table \ref{tab:exp} presents a summary of our conducted experiments, outlining each experiment by its number, the percentage increase in performance achieved by HAAM-RL, and a concise description of the experimental context. The computation of the percentage increase is detailed in Equation \ref{eq:increase}. We compared the number of color changes done by HAAM-RL with that of the heuristic algorithm. 

\begin{equation}
\text{\% Increase} = \frac{\text{HAAM-RL} - \text{Heuristic}}{\text{HAAM-RL}} * 100
\label{eq:increase}
\end{equation}

\begin{figure}[htbp]
    \centerline{\includegraphics[scale=0.8]{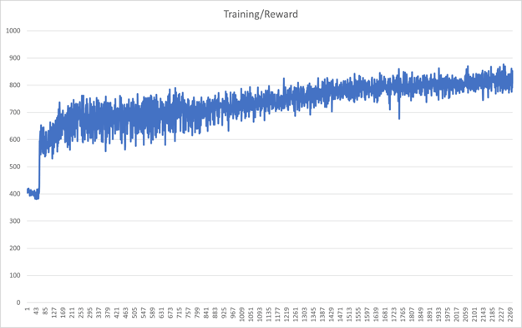}}
    \caption{Training Reward graph}
    \label{fig:FIG8}
\end{figure}

Figure \ref{fig:FIG8} shows the reward graph during the training of the best model; the stable increase in reward is shown. Table \ref{tab:hyperparams} is the set of hyper-parameters that was used during the training. These hyper-parameters were tuned through BakingSoDA and can be modified depending on the problem.
This study demonstrates that the HAAM-RL exhibits superior performance and stability compared to the heuristic algorithms in painting process of the automobile production process. Additionally, it is expected that HAAM-RL will be more effective as the number of vehicles and the complexity of scenario rise.

\section{Discussion and Conclusion}
\label{sec:con}
In this study, we explored various combinations of heuristic algorithms such as Lowest Priority and Color Memory \cite{10.1007/978-3-319-97490-3_36} to apply to efficient training in RL. Some of the future work include the modification of the state representation. The input buffer of the current state only has the incoming vehicle and by having more information about what vehicles will come in the future as input, there is potential for performance improvement. Due to the nature of the problem, we initially utilized only color information when defining the state. However, future experiments should consider incorporating additional information regarding car options and types to further enhance the model \cite{huang2022paint}.

Furthermore, as the problem complexity increases, it may be necessary to encode the state differently and redefine the MDP, as suggested in \cite{10.1007/978-3-319-97490-3_36}. Exploring alternative model-based RL methods \cite{lai2022effective}, \cite{khalid2023sampleefficient}, \cite{WU2024107790} such as Monte Carlo Tree Search (MCTS) \cite{alaniz2018deep}, alongside Proximal Policy Optimization (PPO) \cite{schulman2017proximal} and SAC \cite{haarnoja2018soft} could help uncover differences and potentially achieve higher performance.

\section{Acknowledgments}
The authors would like to express their sincere gratitude to Professor Sara Bysko from Silesian University of Technology for her help with clarification on the heuristic algorithm.

\bibliographystyle{IEEEtran}
\bibliography{references}
\section{Appendix}

\subsection{Experiment}
\label{sec:appendix_exp}

Table \ref{tab:exp_heu} presents the result of heuristic algorithm tested on 30 scenarios, with each scenario having 100 vehicles following the distribution explained in Section \ref{sec:exp}. Table \ref{tab:exp_haam} shows the result of HAAM-RL tested under the same settings. Our methodology outperforms the heuristic algorithm in 29 out of 30 scenarios, achieving an average improvement of 16.25\%.

\begin{table}[htbp]
\centering
\caption{Heuristic Algorithm for 30 Scenarios}
\label{tab:exp_heu}
\begin{tabular}{|c|c|}
\hline
\textbf{Scenario \#} & \textbf{Count} \\
\hline
1 & 36 \\
2 & 36 \\
3 & 36 \\
4 & 36 \\
5 & 32 \\
6 & 33 \\
7 & 38 \\
8 & 37 \\
9 & 35 \\
10 & 32 \\
11 & 39 \\
12 & 30 \\
13 & 33 \\
14 & 34 \\
15 & 34 \\
16 & 30 \\
17 & 33 \\
18 & 34 \\
19 & 34 \\
20 & 34 \\
21 & 34 \\
22 & 35 \\
23 & 36 \\
24 & 36 \\
25 & 34 \\
26 & 34 \\
27 & 34 \\
28 & 35 \\
29 & 36 \\
30 & 37 \\
\hline
\textbf{Average} & \textbf{34.38} \\
\hline
\end{tabular}
\end{table}

\begin{table}[htbp]
\centering
\caption{HAAM-RL with Ensemble Inference Method for 30 Scenarios}
\label{tab:exp_haam}
\begin{tabular}{|c|c|}
\hline
\textbf{Scenario \#} & \textbf{Count} \\
\hline
1 & 27 \\
2 & 29 \\
3 & 29 \\
4 & 32 \\
5 & 29 \\
6 & 34 \\
7 & 34 \\
8 & 31 \\
9 & 34 \\
10 & 26 \\
11 & 27 \\
12 & 26 \\
13 & 29 \\
14 & 29 \\
15 & 27 \\
16 & 29 \\
17 & 31 \\
18 & 28 \\
19 & 31 \\
20 & 31 \\
21 & 31 \\
22 & 31 \\
23 & 32 \\
24 & 29 \\
25 & 27 \\
26 & 31 \\
27 & 32 \\
28 & 33 \\
29 & 29 \\
30 & 29 \\
\hline
\textbf{Average} & \textbf{29.57} \\
\hline
\end{tabular}
\end{table}
\end{document}